\documentclass[journal]{IEEEtran} 

\ifCLASSINFOpdf
\else
\fi

\usepackage{amsmath,graphicx}
\usepackage{tikz}
\usepackage{pgfplots}
\usepackage{pgfplotstable}
\pgfplotsset{compat=newest}
\usepackage{caption}
\usepackage{adjustbox}
\usepackage{booktabs}
\usetikzlibrary{shapes, arrows}
\usetikzlibrary{calc,positioning}
\usetikzlibrary{patterns}
\usepackage{amssymb}
\usepackage{standalone}
\usepackage{pdfpages}
\usepackage{xargs}                      
\usepackage[inline,shortlabels]{enumitem}
\usepackage[colorinlistoftodos,prependcaption,textsize=small, disable]{todonotes} 

\newcommandx{\abst}[2][1=]{\todo[linecolor=blue,backgroundcolor=lime!50,bordercolor=black,#1]{#2}}
\newcommandx{\emre}[2][1=]{\todo[linecolor=red,backgroundcolor=red!10,bordercolor=black,#1]{#2}}
\newcommandx{\tv}[2][1=]{\todo[linecolor=red,backgroundcolor=blue!10,bordercolor=black,#1]{#2}}
\newcommandx{\gb}[2][1=]{\todo[linecolor=red,backgroundcolor=orange!50,bordercolor=black,#1]{#2}}
\newcommandx{\hh}[2][1=]{\todo[linecolor=red,backgroundcolor=yellow!50,bordercolor=black,#1]{#2}}
\newcommandx{\toni}[2][1=]{\todo[linecolor=red,backgroundcolor=green!50,bordercolor=black,#1]{#2}}

\usepackage[normalem]{ulem}
\usepackage{tikz}
\usepackage{caption}

\newcommand{\ra}[1]{\renewcommand{\arraystretch}{#1}}

\usepackage{pdfcomment}

\newcommand\given[1][]{\:#1\vert\:}

\begin{document}
%
\title{Convolutional Recurrent Neural Networks for Polyphonic Sound Event Detection}
%
%
%

\author{Emre~\c{C}ak{\i}r,
        Giambattista~Parascandolo,
        Toni~Heittola,
        Heikki~Huttunen,
        and~Tuomas~Virtanen
\thanks{G. Parascandolo and E. Cakir contributed equally to this work.}
\thanks{The authors are with the Department
of Signal Processing, Tampere University of Technology (TUT), Finland e-mail: emre.cakir@tut.fi.}
\thanks{The research leading to these results has received funding from the European Research Council under the European Union’s H2020 Framework Programme through ERC Grant Agreement 637422 EVERYSOUND.}
\thanks{G.\ Parascandolo has been funded by Google's Faculty Research award.}
\thanks{The authors wish to acknowledge CSC IT Center for Science, Finland, for computational resources.}
\thanks{The paper has a supporting website at

http://www.cs.tut.fi/sgn/arg/taslp2017-crnn-sed/ }
\thanks{Manuscript received July 12, 2016 (revised January 19, 2016).}}

\maketitle

\begin{abstract}
Sound events often occur in unstructured environments where they exhibit wide variations in their frequency content and temporal structure. Convolutional neural networks (CNN) are able to extract higher level features that are invariant to local spectral and temporal variations. Recurrent neural networks (RNNs) are powerful in learning the longer term temporal context in the audio signals. CNNs and RNNs as classifiers have recently shown improved performances over established methods in various sound recognition tasks. We combine these two approaches in a Convolutional Recurrent Neural Network (CRNN) and apply it on a polyphonic sound event detection task. We compare the performance of the proposed CRNN method with CNN, RNN, and other established methods, and observe 
a considerable improvement for four 
different datasets consisting of everyday sound events.    
\end{abstract}


%
\IEEEpeerreviewmaketitle


\section{Introduction}
\IEEEPARstart{I}{n} our daily lives, we encounter a rich variety of sound events such as dog bark, footsteps, glass smash and thunder. Sound event detection (SED), or acoustic event detection, deals with the automatic identification of these sound events. The aim of SED is to detect the onset and offset times for each sound event in an audio recording and associate a textual descriptor, \textit{i.e.}, a label for each of these events. SED has been drawing a surging amount of interest in recent years with applications including audio surveillance~\cite{foggia2015reliable}, healthcare monitoring~\cite{goetze2012acoustic}, urban sound analysis~\cite{salamon2015feature}, multimedia event detection~\cite{wang2016audio} and bird call detection~\cite{stowell2015acoustic}.


In the literature the terminology varies between authors; common terms being sound event \textit{detection}, \textit{recognition}, \textit{tagging} and \textit{classification}. 
Sound events are defined with pre-determined labels called sound event \textit{classes}. In our work, sound event classification, sound event recognition, or sound event tagging, all refer to labeling an audio recording with the sound event classes present, regardless of the onset/offset times. On the other hand, an SED task includes both onset/offset detection for the classes present in the recording and classification within the estimated onset/offset, which is typically the requirement in a real-life scenario.

Sound events often occur in unstructured environments in real-life. Factors such as environmental noise and overlapping sources are present in the unstructured environments and they may introduce a high degree of variation among the sound events from the same sound event class~\cite{dennis2014sound}. Moreover, there can be multiple sound sources that produce sound events belonging to the same class, \textit{e.g.}, a dog bark sound event can be produced from several breeds of dogs with different acoustic characteristics. These factors mainly represent the challenges over SED in real-life situations.



\begin{figure}[!t]
\centering
\includegraphics[width=\linewidth]{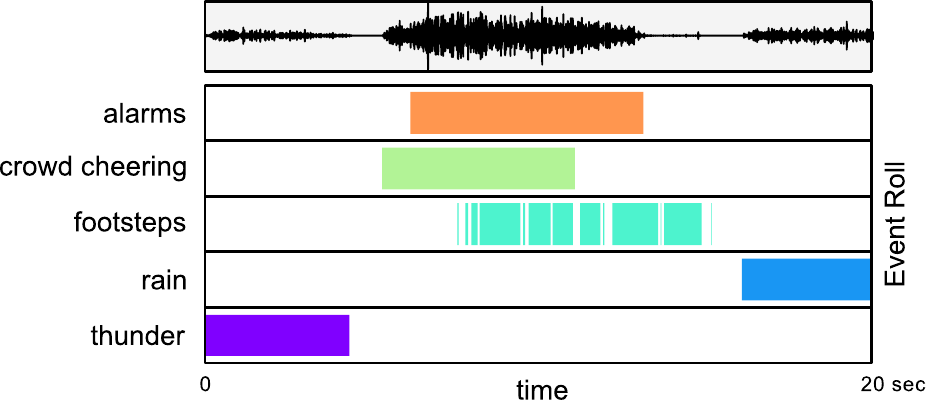} 
\caption{Sound events in a polyphonic recording synthesized with isolated sound event samples. Upper panel: audio waveform, lower panel: sound event class activity annotations.\label{fig:multilabel}}
\end{figure}

SED where at most one simultaneous sound event is detected at a given time instance is called \textit{monophonic} SED. Monophonic SED systems can only detect at most one sound event for any time instance regardless of the number of sound events present. If the aim of the system is to detect all the events happening at a time, this is a drawback concerning the real-life applicability of such systems, because in such a scenario, multiple sound events are very likely to overlap in time. For instance, an audio recording from a busy street may contain footsteps, speech and car horn, all appearing as a mixture of events. An illustration of a similar situation is given in Figure~\ref{fig:multilabel}, where as many as three different sound events appear at the same time in a mixture. A more suitable method for such a real-life scenario is \textit{polyphonic} SED, where multiple overlapping sound events can be detected at any given time instance.

SED can be approached either as \emph{scene-dependent} or \emph{scene-independent}. In the former, the information about the acoustic scene is provided to the system both at training and test time, and a different model can therefore be trained for each scene. In the latter, there is no information about the acoustic scene given to the system.




Previous work on sound events has been mostly focused on sound event classification, where audio clips consisting of sound events are classified. Apart from established classifiers---such as support vector machines~\cite{foggia2015reliable,salamon2015feature}---deep learning methods such as deep belief networks~\cite{Gencoglu}, convolutional neural networks (CNN)~\cite{zhang2015robust, phan2016robust, piczak2015environmental} and recurrent neural networks (RNN)~\cite{wang2016audio,parascandolo2016recurrent} have been recently proposed. Initially, the interest on SED was more focused on monophonic SED. Gaussian mixture model (GMM) - Hidden Markov model (HMM) based modeling---an established method that has been widely used in automatic speech recognition---has been proposed to model individual sound events with Gaussian mixtures and detect each event through HMM states using Viterbi algorithm~\cite{cai2006flexible,mesaros2010acoustic}. With the emergence of more advanced deep learning techniques and publicly available real-life databases that are suitable for the task, polyphonic SED has attracted more interest in recent years. Non-negative matrix factorization (NMF) based source separation~\cite{Mesaros2015_ICASSP} and deep learning based methods (such as feedforward neural networks (FNN)~\cite{cakir2015}, CNN~\cite{cakir2016} and RNN~\cite{parascandolo2016recurrent}) have been shown to perform significantly better compared to established methods such as GMM-HMM for polyphonic SED.

Deep neural networks \cite{lecun2015deep} have recently achieved remarkable success in several domains such as image recognition~\cite{krizhevsky2012imagenet,he2015deep}, speech recognition~\cite{graves2013speech,sainath2015convolutional}, machine translation~\cite{cho2014learning}, even integrating multiple data modalities such as image and text in image captioning~\cite{karpathy2015deep}. 
In most of these domains, deep learning represents the state-of-the-art. 

Feedforward neural networks have been used in monophonic \cite{Gencoglu} and polyphonic SED in real-life environments ~\cite{cakir2015} by processing concatenated input frames from a small time window of the spectrogram. This simple architecture---while vastly improving over established approaches such as GMM-HMMs~\cite{Heittola2013} and NMF source separation based SED~\cite{Heittola2013b,dikmen2013sound}---presents two major shortcomings: (1) it lacks both time and frequency invariance---due to the fixed connections between the input and the hidden units---which would allow to model small variations in the events; (2) temporal context is restricted to short time windows, preventing effective modeling of typically longer events (\emph{e.g.}, rain) and events correlations.

CNNs \cite{lecun1998gradient} can address the former limitation by learning filters that are shifted in both time and frequency \cite{zhang2015robust}, lacking however longer temporal context information. 
Recurrent neural networks (RNNs), which have been successfully applied to automatic speech recognition (ASR) \cite{graves2013speech} and polyphonic SED \cite{parascandolo2016recurrent}, solve the latter shortcoming by integrating information from the earlier time windows, presenting a theoretically unlimited context information. However, RNNs do not easily capture the invariance in the frequency domain, rendering a high-level modeling of the data more difficult.
In order to benefit from both approaches, the two architectures can be combined into a single network with convolutional layers followed by recurrent layers, often referred to as convolutional recurrent neural network (CRNN). Similar approaches combining CNNs and RNNs have been presented recently in ASR \cite{sainath2015convolutional,amodei2016deep,sainath2015learning} and music classification~\cite{choi2016convolutional}.

In this paper we propose the use of multi-label convolutional recurrent neural network for polyphonic, scene-independent sound event detection in real-life recordings. This approach integrates the strengths of both CNNs and RNNs, which have shown excellent performance in acoustic pattern recognition applications
~\cite{wang2016audio,zhang2015robust,phan2016robust,piczak2015environmental}, while overcoming their individual weaknesses. We evaluate the proposed method on three datasets of real-life recordings and compare its performance to FNN, CNN, RNN and GMM baselines. The proposed method is shown to outperform previous sound event detection approaches.

The rest of the paper is organized as follows. In Section~\ref{sec:method}  the problem of polyphonic SED in real-life environments  is described formally and  the CRNN architecture proposed for the task is presented. In Section~\ref{sec:eval} we present the evaluation framework used to measure the performance of the different neural networks architectures. In Section~\ref{sec:results} experimental results, discussions over the results and comparisons with baseline methods are reported. In Section~\ref{sec:concl} we summarize our conclusions from this work. 


\section{Method}
\label{sec:method}

\subsection{Problem formulation}

The aim of polyphonic SED is to temporally locate and label the sound event classes present in a polyphonic audio signal. Polyphonic SED can be formulated in two stages: sound representation and classification. In sound representation stage, frame-level sound features (such as mel band energies and mel frequency cepstral coefficients (MFCC)) are extracted for each time frame $t$ in the audio signal to obtain a feature vector $\mathbf{x}_t \in \mathbb{R}^F$, where $F\in\mathbb{N}$ is the number of features per frame. In the classification stage, the task is to estimate the probabilities $p(\mathbf{y}_t(k) \given \mathbf{x}_t, \boldsymbol{\theta})$ for event classes $k=1,2,\ldots,K$ in frame $t$, where $\boldsymbol{\theta}$ denotes the parameters of the classifier. The event activity probabilities are then binarized by thresholding, \emph{e.g.} over a constant, to obtain event activity predictions $\mathbf{\hat{y}}_t \in \mathbb{R}^K$.

The classifier parameters $\boldsymbol{\theta}$ are trained by supervised learning, and the target outputs $\mathbf{y}_t$ for each frame are obtained from the onset/offset annotations of the sound event classes. If class $k$ is present during frame $t$, $\mathbf{y}_t(k)$ will be set to 1, and 0 otherwise. The trained model will then be used to predict the activity of the sound event classes when the onset/offset annotations are unavailable, as in real-life situations.

For polyphonic SED, the target binary output vector $\mathbf{y}_t$ can have multiple non-zero elements since several classes can be present in the same frame $t$. Therefore, polyphonic SED can be formulated as a multi-label classification problem in which the sound event classes are located by multi-label classification over consecutive time frames. By combining the classification results over consecutive time frames, the onset/offset times for each class can be determined.




Sound events possess temporal characteristics that can be beneficial for SED. Certain sound events can be easily distinguished by their impulsive characteristics (\emph{e.g.}, glass smash), while some sound events typically continue for a long time period (\emph{e.g.} rain). Therefore, classification methods that can preserve the temporal context along the sequential feature vectors are very suitable for SED. For these methods, the input features are presented as a context window matrix $\mathbf{X}_{t:t+T-1}$, where $T \in \mathbb{N}$ is the number of frames that defines the sequence length of the temporal context, and the target output matrix $\mathbf{Y}_{t:t+T-1}$ is composed of the target outputs $\mathbf{y}_t$ from frames $t$ to $t+T-1$. For the sake of simplicity and ease of notation, $\mathbf{X}$ will be used to denote $\mathbf{X}_{t:t+T-1}$---and similarly $\mathbf{Y}$ for $\mathbf{Y}_{t:t+T-1}$--- throughout the rest of the paper.

\subsection{Proposed Method}
\begin{figure}[!t]
\centering
\includegraphics[width=\columnwidth]{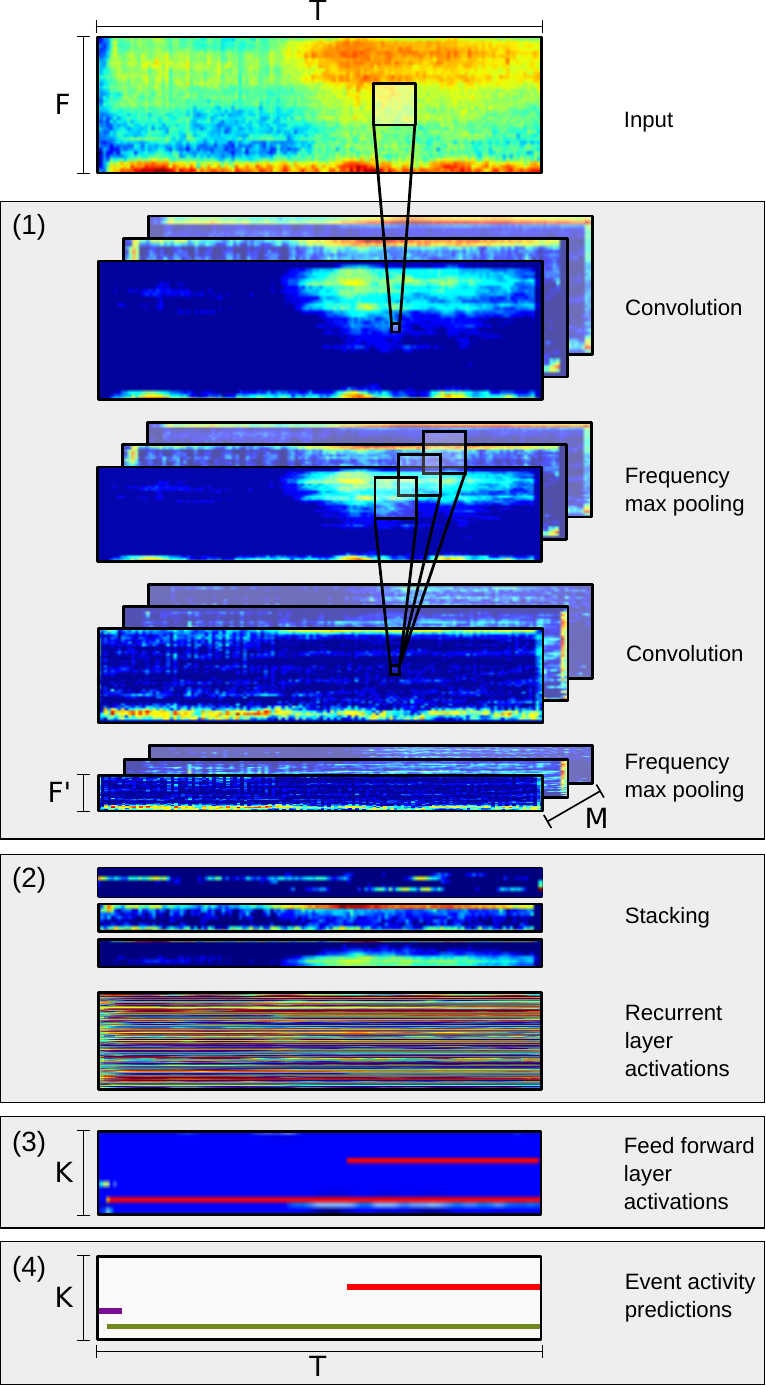}

\caption{Overview of the proposed CRNN method. (1): Multiple convolutional layers with max pooling in frequency axis, (2): The outputs of the last convolutional layer stacked over frequency axis and fed to multiple stacked recurrent layers, (3): feedforward layer as output layer and (4): binarization of event activity probabilities. 
}
\label{fig:TALSP_fig_model}
\end{figure}

The CRNN proposed in this work, depicted in Fig.~\ref{fig:TALSP_fig_model}, consists of four parts: (1) at the top of the architecture, a time-frequency representation of the data (a context window of $F$ log mel band energies over $T$ frames) is fed to $L_c\in\mathbb{N}$ convolutional layers with non-overlapping pooling over frequency axis; (2) the feature maps of the last convolutional layer are stacked over the frequency axis and fed to $L_r\in\mathbb{N}$ recurrent layers; (3) a single feedforward layer with sigmoid activation reads the final recurrent layer outputs and estimates event activity probabilities for each frame and (4) event activity probabilities are binarized by thresholding over a constant to obtain event activity predictions.



In this structure the convolutional layers act as feature extractors, the recurrent layers integrate the extracted features over time thus providing the context information, and finally the feedforward layer produce the activity probabilities for each class. The stack of convolutional, recurrent and feedforward layers is trained jointly through backpropagation. Next, we present the general network architecture in detail for each of the four parts in the proposed method.





\paragraph*{1) Convolutional layers}
Context window of log mel band energies $\textbf{X} \in \mathbb{R}^{F \times T}$ is fed as input to the CNN layers with two-dimensional convolutional filters. For each CNN layer, after passing the feature map outputs through an activation function (rectified linear unit (ReLU) used in this work), non-overlapping max pooling is used to reduce the dimensionality of the data and to provide more frequency invariance. As depicted in Fig. \ref{fig:TALSP_fig_model}, the time dimension is maintained intact (\textit{i.e.} does not shrink) by computing the max pooling operation in the frequency dimension only---as done in \cite{sainath2015convolutional,sigtia2016end}---and by zero-padding the inputs to the convolutional layers (also known as \textit{same} convolution). This is done in order to preserve alignment between each target output vector $\mathbf{y}_t$ and hidden activations $\mathbf{h}_t$.


After $L_c$ convolutional layers, the output of the CNN is a tensor $\mathcal{H} \in \mathbb{R}^{M\times F^\prime \times T}$, where $M$ is the number of feature maps for the last CNN layer, and $F^\prime $ is the number of frequency bands remaining after several pooling operations through CNN layers.





\paragraph*{2) Recurrent layers}
After stacking the feature map outputs over the frequency axis, the CNN output $\mathbf{H}\in \mathbb{R}^{(M\cdot F^\prime)\times T}$ for layer $L_c$ is fed to the RNN as a sequence of frames $\mathbf{h}^{L_c}_t$. The RNN part consists of $L_r$ stacked recurrent layers each computing and outputting a hidden vector $\mathbf{h}_t$ for each frame as 

\begin{equation}
\begin{aligned}
\mathbf{h}^{L_c+1}_t &= \mathcal{F}(\mathbf{h}^{L_c}_t, \mathbf{h}^{L_c+1}_{t-1}) \\
\mathbf{h}^{L_c+2}_t &= \mathcal{F}(\mathbf{h}^{L_c+1}_t, \mathbf{h}^{L_c+2}_{t-1}) \\
&\mathrel{\makebox[\widthof{=}]{\vdots}} \\
\mathbf{h}^{L_c+L_r}_t &= \mathcal{F}(\mathbf{h}^{L_c+L_r-1}_t, \mathbf{h}^{L_c+L_r}_{t-1}) 
\end{aligned}
\end{equation}
The function $\mathcal{F}$, which can represent a long short term memory (LSTM) unit~\cite{hochreiter1997long} or gated recurrent unit (GRU)~\cite{cho2014properties}, has two inputs: The output of the current frame of the previous layer (\emph{e.g.,} $\mathbf{h}^{L_c}_t$), and the output of the previous frame of the current layer (\emph{e.g.,} $\mathbf{h}^{L_c+1}_{t-1}$).




\paragraph*{3) Feedforward layer}




recurrent layers are followed by a single feedforward layer which will be used as the output layer of the network. The feedforward layer outputs are obtained from the last recurrent layer activations $\mathbf{h}^{L_c+L_r}_t$ as  
\begin{equation}
\mathbf{h}^{L_c+L_r+1}_t = \mathcal{G}(\mathbf{h}^{L_c + L_r}_t),
\end{equation}
where $\mathcal{G}$ represents a feedforward layer with sigmoid activation. Feedforward layer applies the same set of weights for the features extracted from each frame.


\paragraph*{4) Binarization}
The outputs $\mathbf{h}^{L_c+L_r+1}_t$ of the feedforward layer are used as the event activity probabilities for each class $k =1,2,...K$ as 

\begin{equation}
p(\mathbf{y}_t(k) \given \mathbf{x}_{0:t}, \boldsymbol{\theta}) = \mathbf{h}^{L_c+L_r+1}_t
\end{equation}
where $K$ is the number of classes and $\boldsymbol{\theta}$ represents the parameters of all the layers of the network combined.
Finally, event activity predictions $\mathbf{\hat{y}}_t$ are obtained by thresholding the probabilities over a constant $C\in(0,1)$ as
\begin{equation}
	\mathbf{\hat{y}}_t(k) = \begin{cases}
	1, &p(\mathbf{y}_t(k) \given \mathbf{x}_{0:t}, \boldsymbol{\theta}) \geq C \\
    0, &\text{otherwise}
    \end{cases}
\end{equation}



\paragraph*{Regularization}
In order to reduce overfitting, we experimented with dropout~\cite{srivastava2014dropout} regularization in the network, which has proven to be extremely effective in several deep learning applications~\cite{krizhevsky2012imagenet}. The basic idea behind dropout is to temporarily remove at training time a certain portion of hidden units from the network, with the dropped units being randomly chosen at each iteration. This reduces units co-adaptation, approximates model averaging \cite{srivastava2014dropout}, and can be seen as a form of data augmentation without domain knowledge.
For the recurrent layers we adopted the dropout proposed in \cite{gal2015theoretically}, where the choice of dropped units is kept constant along a sequence.

To speed up the training phase we train our networks with batch normalization layers \cite{ioffe2015batch} after every convolutional or fully connected layer. Batch normalization reduces the internal covariate shift---\emph{i.e.}, the distribution of network activations during training---by normalizing a layer output to zero mean and unit variance, using approximate statistics computed on the training mini-batch.

\paragraph*{Comparison to other CRNN architectures}
The CRNN configuration used in this work has several points of similarity with the network presented in~\cite{sainath2015convolutional} for speech recognition. The main differences are the following:
\begin{enumerate*}[(i)]
\item
We do not use any linear projection layer, neither at the end of the CNN part of the CRNN, nor after each recurrent layer. 
\item We use 5x5 kernels in all of our convolutional layers, compared to the 9x9 and 4x3 filters for the first and second layer respectively. 
\item Our architecture has also more convolutional layers (up to 4 instead of 2) and recurrent layers (up to 3 instead of 2). 
\item We use GRU instead of LSTM. 
\item We use much longer sequences, up to thousands of steps, compared to 20 steps in \cite{sainath2015convolutional}. While very long term context is not helpful in speech processing, since words and utterances are quite short in time, in SED there are several events that span over several seconds. 
\item For the experiments on CHiME-Home dataset we incorporate a new max pooling layer (only on time domain) before the output layer. Therefore, if we have $N$ mid-level features for $T$ frames of a context window, we end up with $N$ features for the whole context window to be fed to the output layer.
\end{enumerate*}
\paragraph*{CNNs and RNNs}
It is possible to see CNNs and RNNs as specific instances of the CRNN architecture presented in this section: a CNN is a CRNN with zero recurrent layers, and an RNN is a CRNN with zero convolutional layers. In order to assess the benefits of using CRNNs compared to CNNs or RNNs alone, in Section III we directly compare the three architectures by removing the recurrent or convolutional layer, \emph{i.e.}, CNNs and RNNs respectively.





\section{Evaluation}
\label{sec:eval}

In order to test the proposed method, we run a series of experiments on four different datasets. We evaluate the results by comparing the system outputs to the annotated references. Since we are approaching the task as scene-independent, on each dataset we train a single model regardless of the presence of different acoustic scenes.


\subsection{Datasets and Settings}

We evaluate the proposed method on four datasets, one of which is artificially generated as mixtures of isolated sound events, and three are recorded from real-life environments. 

While an evaluation performed on real audio data would be ideal, human annotations tend to be somewhat subjective, especially when precise onset and offset are required for overlapping events. For this reason we create our own synthetic dataset---from here onwards referred to as \textit{TUT Sound Events Synthetic 2016} --- where we use frame energy based automatic annotation of sound events.

In order to evaluate the proposed method in real-life conditions, we use \textit{TUT Sound Events 2009}. This proprietary dataset contains real-life recordings from 10 different scenes and has been used in many previous works. We also compute and show results on the \textit{TUT Sound Events 2016 development} and \textit{CHiME-Home} dataset, which were used as part of DCASE2016 challenge~\footnote{http://www.cs.tut.fi/sgn/arg/dcase2016/}.




\paragraph{TUT Sound Events Synthetic 2016 (TUT-SED Synthetic 2016)} 

The primary evaluation dataset consists of synthetic mixtures created by mixing isolated sound events from 16 sound event classes. Polyphonic mixture were created by mixing 994 sound event samples. From the 100 mixtures created, 60\% are used for training, 20\% for testing and 20\% for validation. The total length of the data is 566 minutes. Different instances of the sound events are used to synthesize the training, validation and test partitions. Mixtures were created by randomly selecting event instance and from it, randomly, a segment of length 3-15 seconds. Mixtures do not contain any additional background noise. Dataset creation procedure explanation and metadata can be found in the supporting website for the paper\footnote{http://www.cs.tut.fi/sgn/arg/taslp2017-crnn-sed/tut-sed-synthetic-2016}.

\paragraph{TUT Sound Events 2009 (TUT-SED 2009)}
This dataset, first presented in \cite{heittola2010audio}, consists of 8 to 14 binaural recordings from 10 real-life scenes. Each recording is 10 to 30 minutes long, for a total of 1133 minutes. The 10 scenes are: basketball game, beach, inside a bus, inside a car, hallway, office, restaurant, shop, street and stadium with track and field events. A total of 61 classes were defined, including (wind, yelling, car, shoe squeaks, etc.) and one extra class for unknown or rare events. The average number of events active at the same time is 2.53. Event activity annotations were done manually, which introduces a degree of subjectivity.
The database has a five-fold cross-validation setup with training, validation and test set split, each consisting of about 60\%, 20\% and 20\% of the data respectively from each scene. The dataset unfortunately can not be made public due to licensing issues, however three $\sim10$ minutes samples from the dataset are available at
\footnote{http://arg.cs.tut.fi/demo/CASAbrowser/}.




\paragraph{TUT Sound Events 2016 development  (TUT-SED 2016)}



This dataset 
consists of recordings from two real-life scenes: residential area and home \cite{Mesaros2016_EUSIPCO}. 
The recordings are captured each in a different location (\emph{i.e.}, different streets, different homes) leading to a large variability on active sound event classes between recordings. For each location, a 3-5 minute long binaural audio recording is provided, adding up to 78 minutes of audio. The recordings have been manually annotated. In total, there are seven annotated sound event classes for residential area recordings and 11 annotated sound event classes for home recordings. The dataset and metadata is available through \footnote{http://www.cs.tut.fi/sgn/arg/taslp2017-crnn-sed/\#tut-sed-2016} and \footnote{https://zenodo.org/record/45759\#.WBoUGrPIbRY}.

The four-fold cross-validation setup published along with the dataset~\cite{Mesaros2016_EUSIPCO} is used in the evaluations. 
Twenty percent of the training set recordings are assigned for validation in the training stage of the neural networks. 
Since in this work we investigate scene-independent SED, we discard the information about the scene, contrary to the DCASE2016 challenge setup. Therefore, instead of training a separate classifier for each scene, we train a single classifier to be used in all scenes. 
In TUT-SED 2009 all audio material for a scene was recorded in a single location, whereas TUT-SED 2016 contains multiple locations per scene.


\paragraph{CHiME-Home}
CHiME-Home dataset~\cite{foster2015chime} consists of 4-second audio chunks from home environments. The annotations are based on seven sound classes, namely child speech, adult male speech, adult female speech, video game / TV, percussive sounds, broadband noise and other identifiable sounds. In this work, we use the same, \textit{refined} setup of CHiME-Home as it is used in audio tagging task in DCASE2016 challenge~\cite{DCASE2016}, namely 1946 chunks for development (in four folds) and 846 chunks for evaluation.

The main difference between this dataset and the previous three is that the annotations are made per chunk instead of per frame. Each chunk is annotated with one or multiple labels. In order to adapt our architecture to the lack of frame-level annotations, we simply add a temporal max-pooling layer---that pools the predictions over time---before the output layer for FNN, CNN, RNN and CRNN. CHiME-Home dataset is available at \footnote{https://archive.org/details/chime-home}.


\subsection{Evaluation Metrics}
\label{subsec:evalmet}

\begin{table*}[t]
\centering
\ra{1.1}
\caption{Final hyperparameters used for the evaluation based on the validation results from the hyperparameter grid search.}
\label{tab:params}\normalsize
\resizebox{\linewidth}{!}{%
\begin{tabular}{ @{} l *{3}{c c c|} *{1}{c c c}}
\toprule
& \multicolumn{3}{c|}{TUT-SED Synthetic 2016} & \multicolumn{3}{c|}{TUT-SED 2009} & \multicolumn{3}{c|}{TUT-SED 2016} & \multicolumn{3}{c}{CHiME-Home} \\
								&CNN	&RNN	&CRNN	&CNN	&RNN	&CRNN	&CNN	&RNN	&CRNN	&CNN	&RNN	&CRNN \\
\midrule
\# CNN layers					&3		& -		&3		&3		& -		&3		&3		& -		&3		&3		& -		&4	\\
pool size 						&(2,2,2)& -		&(5,4,2)&(5,4,2)& -		&(5,4,2)&(5,4,2)& -		&(2,2,2)&(5,4,2)& -		&(2,2,2,1)	\\
\# RNN layers					& -		&3		&1		& -		&3		&1		& -		&3		&3		& -		&2		&1	\\
\# FNN layers					&3		&2		& -		&1		&4		& -		&1		&4		& -		&1		&1		& -	\\
\# feature maps/hidden units 	&256		&512	&256	&256	&256	&256	&256	&256	&96	&256	&256	&256	\\
sequence length	(s)				&2.56	&5.12	&20.48	&2.56	&20.48	&20.48	&2.56	&20.48	&2.56	&4		&4		&4	\\
\# Parameters					&3.7M	&4.5M	&3.6M	&3.4M	&1.3M	&3.7M	&3.4M	&1.3M	&743K	&3.6M	&690K	&6.1M	\\
\bottomrule
\end{tabular}
}
\end{table*}

In this work, segment-based evaluation metrics are used. The segment lengths used in this work are (1): a single time frame (40 ms in this work) and (2): a one-second segment. The segment length for each metric is annotated with the subscript (\textit{e.g.},  $F1_\textnormal{frm}$ and $F1_\textnormal{1sec}$).

Segment-based F1 score calculated in a single time frame ($F1_\textnormal{frm}$) is used as the primary evaluation metric~\cite{mesaros2016metrics}. For each segment in the test set, intermediate statistics, \textit{i.e.}, the number of true positive (\textit{TP}), false positive (\textit{FP}) and false negative (\textit{FN}) entries, are calculated as follows. If an event
\begin{itemize}
\item{is detected in one of the frames inside a segment and it is also present in the same segment of the annotated data, that event is regarded as \textit{TP}.}
\item{is \textit{not} detected in any of the frames inside a segment but it is present in the same segment of the annotated data, that event is regarded as \textit{FN}.}
\item{is detected in one of the frames inside a segment but it is \textit{not} present in the same segment of the annotated data, that event is regarded as \textit{FP}.}
\end{itemize}
These intermediate statistics are accumulated over the test data and then over the folds. This way, each active instance per evaluated segment has equal influence on the evaluation score. This calculation method is referred to as micro-averaging, and is the recommended method for evaluation of classifier \cite{Forman10}. Precision ($P$) and recall ($R$) are calculated from the accumulated intermediate statistics as 
\begin{equation}
P = \frac{\textit{TP}}{\textit{TP}+\textit{FP}} 
\qquad
R = \frac{\textit{TP}}{\textit{TP}+\textit{FN}} 
\end{equation}
These two metrics are finally combined as their harmonic mean, \textit{F1 score}, which can be formulated as 
\begin{equation}
	\label{eq:FmeasureEq}
    \textit{F1} = \frac{2 \cdot P \cdot R}{P + R}
\end{equation}
More detailed and visualized explanation of segment-based F1 score in multi label setting can be found in~\cite{mesaros2016metrics}. 

The second evaluation metric is segment-based error rate as proposed in~\cite{mesaros2016metrics}. For error rate, intermediate statistics, \textit{i.e.}, the number of substitutions~(\textbf{s}), insertions~(\textbf{i}), deletions~(\textbf{d}) and active classes from annotations~(\textbf{a}) are calculated per segment as explained in detail in~\cite{mesaros2016metrics}. Then, the total error rate is calculated as
\begin{equation}
\label{eq:ER}
ER = \frac{\sum\limits_{t=1}^N\mathbf{s}_t +\sum\limits_{t=1}^N\mathbf{i}_t +\sum\limits_{t=1}^N\mathbf{d}_t}{\sum\limits_{t=1}^R\mathbf{a}_t}
\end{equation}
where subscript $t$ represents segment index and N is the total number of segments.

Both evaluation metrics are calculated from the accumulated sum for their corresponding intermediate statistics over the segments of the whole test set. If there are multiple scenes in the dataset, evaluation metrics are calculated for each scene separately and then the results are presented as the average across the scenes. 



The main metric used in previous works~\cite{parascandolo2016recurrent,Mesaros2015_ICASSP,cakir2015} on TUT-SED 2009 dataset differs from the F1 score calculation used in this paper. In previous works, F1 score was computed in each segment, then averaged along segments for each scene, and finally averaged across scene scores, instead of accumulating intermediate statistics. This leads to measurement bias under high class imbalance between the classes and also between folds. However, in order to give a comprehensive comparison of our proposed method with previous works on this dataset, we also report the results with this legacy F1 score in Section~\ref{subsec:casa}.

For CHiME-Home dataset, equal error rate (EER) has been used as the evaluation metric in order to compare the results with DCASE2016 challenge submissions, where EER has been the main evaluation metric.

\subsection{Baselines}
For this work, we compare the proposed method with two recent approaches: the Gaussian mixture model (GMM) of~\cite{Mesaros2016_EUSIPCO} and the feedforward neural network model (FNN) from~\cite{cakir2015}. GMM has been chosen as a baseline method since it is an established generative modeling method used in many sound recognition tasks~\cite{cai2006flexible,mesaros2010acoustic,cheng2003semantic}. In parallel with the recent surge of deep learning techniques in pattern recognition, FNNs have been shown to vastly outperform GMM based methods in SED~\cite{cakir2015}. Moreover, this FNN architecture represents a straightforward deep learning method that can be used as a baseline for more complex architectures such as CNN, RNN and the proposed CRNN.

\paragraph*{GMM} The first baseline system is based on a binary frame-classification approach, where for each sound event class a binary classifier is set up \cite{Mesaros2016_EUSIPCO}. Each binary classifier consists of a positive class model and a negative class model. The positive class model is trained using the audio segments annotated as belonging to the modeled event class, and a negative class model is trained using the rest of the audio. The system uses MFCCs as features and a GMM-based classifier. MFCCs are calculated using 40 ms frames with Hamming window and 50\% overlap and 40 mel bands. The first 20 static coefficients are kept, and delta and acceleration coefficients are calculated using a window length of 9 frames. The 0th order static coefficient is excluded, resulting in a frame-based feature vector of dimension 59. For each sound event, a positive model and a negative model are trained. The models are trained using expectation-maximization algorithm, using k-means algorithm to initialize the training process and diagonal covariance matrices. The number of parameters for GMM baseline is $3808*K$, where $K$ is the number of classes. In the detection stage, the decision is based on the likelihood ratio between the positive and negative models for each individual sound class event, with a sliding window of one second. The system is used as a baseline in the DCASE2016 challenge \cite{DCASE2016_baseline}, however, in this study the system is used as scene-independent to match the setting of the other methods presented.

\paragraph*{FNN} 
The second baseline system is a deep multi-label FNN with temporal context~\cite{cakir2015}. As the sound features, 40 log mel band energy features are extracted for each 40 ms time frame with 50\% overlap. For the input, consecutive feature vectors are stacked in five vector blocks, resulting in a 100 ms context window. As the hidden layers, two feedforward layers of 1600 hidden units with maxout activation~\cite{goodfellow2013maxout} with pool size of 2 units are used. For the output layer, a feedforward layer of $K$ units with sigmoid activation is used to obtain event activity probabilities per context window, where $K$ is the number of classes. The sliding window post-processing of the event activity probabilities in~\cite{cakir2015} has not been implemented for the baseline experiments in order to make a fair comparison based on classifier architecture for different deep learning methods. The number of parameters in the baseline FNN model is around 1.6 million.


\subsection{Experiments set-up}

\begin{table*}[!t]
\centering
\ra{1.1}
\caption{F1 score and error rate results for single frame segments ($F1_{\textnormal{frm}}$ and $ER_{\textnormal{frm}}$) and one second segments ($F1_{\textnormal{1sec}}$ and $ER_{\textnormal{1sec}}$). Bold face indicates the best performing method for the given metric. }
\label{tab:general} \normalsize
\resizebox{\linewidth}{!}{%
\begin{tabular}{ @{} l *{2}{c c c c|} *{1}{c c c c}} 
\toprule
 & \multicolumn{4}{c|}{TUT-SED Synthetic 2016} &\multicolumn{4}{c|}{TUT-SED 2009} & \multicolumn{4}{c}{TUT-SED 2016}\\

Method & $F1_{\textnormal{frm}}$ & $ER_{\textnormal{frm}}$ & $F1_{\textnormal{1sec}}$ & $ER_{\textnormal{1sec}}$ & $F1_{\textnormal{frm}}$ & $ER_{\textnormal{frm}}$ & $F1_{\textnormal{1sec}}$ & $ER_{\textnormal{1sec}}$ & $F1_{\textnormal{frm}}$ & $ER_{\textnormal{frm}}$ & $F1_{\textnormal{1sec}}$ & $ER_{\textnormal{1sec}}$\\
\midrule
GMM~\cite{Mesaros2016_EUSIPCO} 	& 40.5 &0.78 & 45.3 & 0.72	&33.0	&1.34	&34.1	&1.60	&14.1		&1.12	&17.9	&1.13 \\
FNN~\cite{cakir2015}    		& 49.2$\pm$0.8  &0.68$\pm$0.02 & 50.2$\pm$1.4 & 1.1$\pm$0.1	&60.9$\pm$0.4	&0.56$\pm$0.01	&57.1$\pm$0.2	&1.1$\pm$0.01	&26.7$\pm$1.4		&\textbf{0.99$\pm$0.03}	&\textbf{32.5$\pm$1.2}	&1.32$\pm$0.06 \\
CNN    							& 59.8$\pm$0.9  &0.56$\pm$0.01 & 59.9$\pm$1.2 & 0.78$\pm$0.08	&64.8$\pm$0.2	&0.50$\pm$0.0		&63.2$\pm$0.5	&0.75$\pm$0.02	&23.0$\pm$2.6	&1.02$\pm$0.06	&26.4$\pm$1.9	&1.09$\pm$0.06 \\
RNN    							& 52.8$\pm$1.5  &0.6$\pm$0.02 & 57.1$\pm$0.9 & 0.64$\pm$0.01	&62.4$\pm$1.0	&0.52$\pm$0.01	&61.8$\pm$0.8	&0.55$\pm$0.01	&\textbf{27.6$\pm$1.8}	&1.04$\pm$0.02	&29.7$\pm$1.4	&1.10$\pm$0.04 \\
{\bf CRNN}  					& {\bf 66.4$\pm$0.6} &{\bf0.48$\pm$0.01} & {\bf68.7$\pm$0.7} & {\bf0.47$\pm$0.01}	&\textbf{69.7$\pm$0.4}	&\textbf{0.45$\pm$0.0}	&\textbf{69.3$\pm$0.2}	&\textbf{0.48$\pm$0.0}	&\textbf{27.5$\pm$2.6} 	&\textbf{0.98$\pm$0.04} 	&30.3$\pm$1.7	&\textbf{0.95$\pm$0.02}   \\

\bottomrule
\end{tabular}
}
\end{table*}

\paragraph*{Preprocessing}
For all neural networks (FNN, CNN, RNN and CRNN) we use log mel band energies as acoustic features. We first compute short-time Fourier transform (STFT) of the recordings in 40 ms frames with 50\% overlap, then compute mel band energies through mel filterbank with 40 bands spanning 0 to 22050 Hz, which is the Nyquist rate. After computing the logarithm of the mel band energies, each energy band is normalized by subtracting its mean and dividing by its standard deviation computed over the training set. 
The normalized log mel band energies are finally split into sequences. During training we use overlapped sequences, \textit{i.e.} we sample the sub-sequences with a different starting point at every epoch, by moving the starting index by a fixed amount that is not a factor of the sequence length (73 in our experiments). The stride is not equal to 1 in order to have effectively different sub-sequences from one training epoch to the next one. For validation and test data we do not use any overlap.

While finer frequency resolution or different representations could improve the accuracy, our main goal is to compare the architectures. We opted for this setting as it was recently used with very good performance in several works on SED \cite{parascandolo2016recurrent,cakir2015}.
\paragraph*{Neural network configurations}
Since the size of the dataset usually affects the optimal network architecture, we do a hyperparameter search by running a series of experiments over predetermined ranges. We select for each network architecture the hyperparameter configuration that leads to the best results on the validation set, and use this architecture to compute the results on the test set. 

For TUT-SED Synthetic 2016  and CHiME-Home datasets, we run a hyperparameter grid search on the number of CNN feature maps and RNN hidden units \{96, 256\} (set to the same value); the number of recurrent layers \{1, 2, 3\}; and the number of CNN layers \{1, 2, 3 ,4\} with the following frequency max pooling arrangements after each convolutional layer \{(4), (2, 2), (4, 2), (8, 5), (2, 2, 2), (5, 4, 2), (2, 2, 2, 1), (5, 2, 2, 2)\}.  Here, the numbers denote the number of frequency bands at each max pooling step; \emph{e.g.,} the configuration (5, 4, 2) pools the original 40 bands to one band in three stages: 40 bands $\rightarrow$ 8 bands $\rightarrow$ 2 bands $\rightarrow$ 1 band. 

\begin{figure}[!t]
\includegraphics[width=\linewidth]{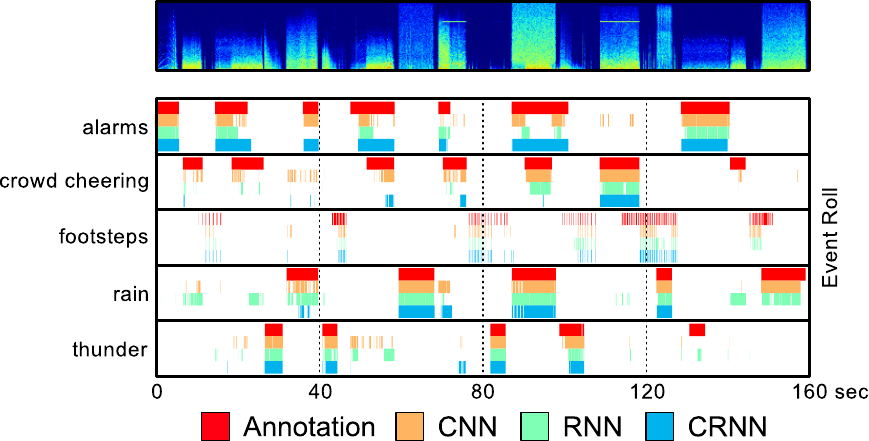} 
\caption{Annotations and event activity predictions for CNN, RNN and CRNN over a mixture from TUT-SED Synthetic 2016. For clarity, the classes that are not present in the mixture are omitted.}
\label{fig:outputs}
\end{figure}

All networks have batch normalization layers after convolutional layers and dropout rate 0.25, which were found to be helpful in preliminary experiments. The output layer consists of a node for each class and has the sigmoid as activation function. In convolutional layers we use filters with shape $(5,5)$; in recurrent layers we opted for GRU, since preliminary experiments using LSTM yielded similar results and GRU units have a smaller number of parameters. The weights are initialized according to the scheme proposed in \cite{he2015delving}. Binary cross-entropy is set as the loss function, and all networks are trained with Adam \cite{kingma2014adam} as gradient descent optimizer, with the default parameters proposed in the original paper. 

To evaluate the effect of having both convolutional and recurrent layers in the same architecture, we compare the CRNN with CNNs and RNNs alone. For both CNN and RNN we run the same hyperparameter optimization procedure described for CRNN, replacing recurrent layers with feedforward layers for CNNs, and removing convolutional layers for RNNs while adding feedforward layers before the output layer. This allows for a fair comparison, providing the possibility of having equally deep networks for all three architectures.

After this first optimization process, we use the best CRNNs, CNNs and RNNs to separately test the effect of varying other hyperparameters. More specifically we investigate how performance is affected by variation of the CNN filter shapes and the sequence length. For the CRNN we test filter shapes in the set~\{(3,3), (5,5), (11,11), (1,5), (5,1), (3,11), (11,3)\}, where ($\ast,\ast$) represents the filter lengths in frequency and time axes, respectively. For CRNN and RNN, we test shorter and longer sequences than the initial value of 128 frames, experimenting in the range \{8, 32, 128, 256, 512, 1024, 2048\} frames, which correspond to \{0.16, 0.64, 2.56, 5.12, 10.24, 20.48, 40.96\} seconds respectively. We finally use the hyperparameters that provide the highest validation scores as our final CRNN, CNN and RNN models.

For the other two datasets (TUT-SED 2009 and TUT-SED 2016) we select a group of best performing model configurations on validation data from TUT-SED Synthetic 2016 experiments and to account for the different amount of data we run another smaller hyperparameter search, varying the amount of dropout and the sequence length. Again, we then select the best performing networks on the validation score to compute the test results. The hyperparameters used in the evaluation for all three datasets is presented in Table~\ref{tab:params}.

The event activity probabilities are thresholded at $C=0.5$, in order to obtain the binary activity matrix used to compute the reference metrics based on the ground truth. All networks are trained until overfitting starts to arise: as a criterion we use early stopping on the validation metric, halting the training if the score is not improving for more than 100 epochs and reverting the weights to the values that best performed on validation. 



\begin{table}[!t]
\centering
\ra{1.1}
\caption{$F1_\textnormal{frm}$ for CNN, RNN and CRNN for each class in TUT-SED Synthetic 2016. }
\label{tab:class2} \normalsize
\resizebox{\columnwidth}{!}{%
\begin{tabular}{ @{} l *{5}{c}} 
\toprule
Class & avg. (secs) &	total (secs)	& CNN & RNN  & CRNN  \\
\midrule
glass smash		&1.2	&621	&\textbf{57$\pm$8.6}&	48$\pm$2.0&	\textbf{54$\pm$6.7}\\
gun shot		&1.7	&534	&53$\pm$5.9&	64$\pm$2.3&	\textbf{73$\pm$1.8}\\
cat meowing		&2.1	&941	&\textbf{37$\pm$4.6}&	29$\pm$4.5&	\textbf{42$\pm$3.9}\\
dog barking		&5.0	&716	&\textbf{69$\pm$3.3}&	51$\pm$2.5&	\textbf{73$\pm$3.1}\\
thunder			&5.9	&3007	&55$\pm$3.3&	46$\pm$2.2&	\textbf{63$\pm$1.9}\\
bird singing 	&6.1	&2298	&44$\pm$1.2&	41$\pm$3.1&	\textbf{53$\pm$2.3}\\
horse walk		&6.4	&1614	&\textbf{46$\pm$2.1}&	39$\pm$2.7&	\textbf{45$\pm$2.4}\\
baby crying		&6.9	&2007	&46$\pm$5.7&	46$\pm$1.1&	\textbf{59$\pm$3.0}\\
motorcycle		&7.0	&3691	&\textbf{47$\pm$3.1}&	\textbf{44$\pm$2.2}&	\textbf{47$\pm$2.7}\\
footsteps		&7.1	&1173	&41$\pm$2.0&	34$\pm$1.2&	\textbf{47$\pm$1.7}\\
crowd applause	&7.3	&3278	&\textbf{68$\pm$1.8}&	57$\pm$1.5&	\textbf{71$\pm$0.6}\\
bus				&7.8	&3464	&60$\pm$2.0&	55$\pm$2.5&	\textbf{66$\pm$2.4}\\
mixer			&7.9	&4020	&62$\pm$5.6&	57$\pm$6.4&	\textbf{82$\pm$2.7}\\
crowd cheering	&8.1	&4825	&72$\pm$2.9&	64$\pm$2.7&	\textbf{77$\pm$1.1}\\
alarms 			&8.2	&4405	&\textbf{64$\pm$2.2}&	50$\pm$5.3&	\textbf{66$\pm$2.9}\\
rain			&8.2	&3975	&\textbf{71$\pm$2.0}&	59$\pm$2.6&	\textbf{72$\pm$1.9}\\

\bottomrule
\end{tabular}
}
\end{table}

For feature extraction, the Python library Librosa ~\cite{mcfee2015librosa} has been used in this work. For classifier implementations, deep learning package Keras (version 1.1.0)~\cite{Keras} is used with Theano (version 0.8.2) as backend~\cite{team2016theano}. The networks are trained on NVIDIA Tesla K40t and K80 GPUs. 

\section{Results}
\label{sec:results}

In this section, we present results for all the datasets and experiments described in Section \ref{sec:eval}. The evaluation of CNN, RNN and CRNN methods are conducted using the hyperparameters given in Table~\ref{tab:params}. All the reported results are computed on the test sets. Unless otherwise stated, we run each neural network based experiment ten times with different random seeds (five times for TUT-SED 2009) to reflect the effect of random weight initialization. We provide the mean and the standard deviation of these experiments in this section. Best performing method is highlighted with bold face in the tables of this section. The methods whose best performance among the ten runs is within one standard deviation of the best performing method is also highlighted with bold face.

The main results with the best performing (based on the validation data) CRNN, CNN, RNN, and the GMM and FNN baselines are reported in Table \ref{tab:general}. 
Results are calculated according to the description in Section~ \ref{subsec:evalmet} where each event instance irrespective of the class is taken into account in equal manner. 
As shown in the table, the CRNNs consistently outperforms CNNs, RNNs and the two baseline methods on all three datasets for the main metric. 

\subsection{TUT Sound Events Synthetic 2016}




As presented in Table~\ref{tab:general}, CRNN improved by absolute 6.6\%
and 13.6\%
on frame-based F1 compared to CNN and RNN respectively for TUT-SED synthetic 2016 dataset. Considering the number of parameters used for each method (see Table~\ref{tab:params}), the performance of CRNN indicates an architectural advantage compared to CNN and RNN methods. All the four deep learning based methods outperform the baseline GMM method. 
As claimed in~\cite{hinton2012deep}, this may be due to the capability of deep learning methods to use different subsets of hidden units to model different sound events simultaneously. An example mixture from TUT-SED Synthetic 2016 test set is presented in Figure~\ref{fig:outputs} with annotations and event activity predictions from CNN, RNN and CRNN.  

\begin{table}[!t] 
\centering
\ra{1.1}
\caption{$F1_{\textnormal{frm}}$ for accuracy vs. convolution filter shape for TUT-SED Synthetic 2016 dataset. ($\ast,\ast$) represents filter lengths in frequency and time axis, respectively.}
\label{tab:filt} \normalsize
\resizebox{\columnwidth}{!}{%
\begin{tabular}{ @{} l *{7}{c}}  
\toprule
Filter shape &(3,3)	& (5,5)  &(11,11) &(1,5) &(5,1) &(3,11) &(11,3)   \\
\midrule
$F1_{\textnormal{frm}}$  		& 67.2	& \textbf{68.3}  & 62.6 & 28.5 & 60.6& 67.4 & 61.2 \\
\bottomrule
\end{tabular}
} 
\end{table} 

\begin{figure}[!t]
\includegraphics[width=\columnwidth]{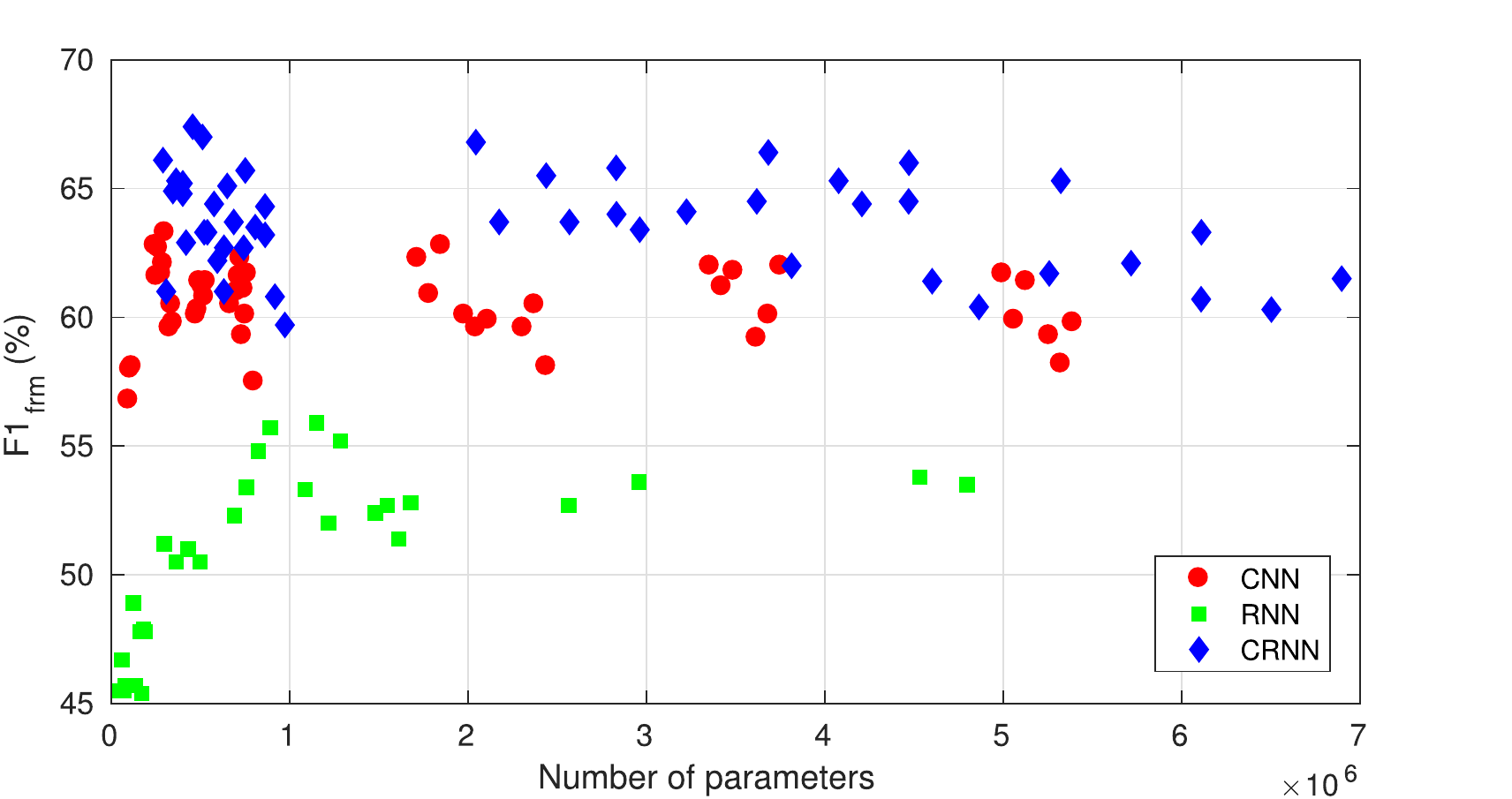} 
\caption{Number of parameters vs. accuracy for CNN, RNN and CRNN.}

\label{fig:num_params}
\end{figure}

\subsubsection{Class-wise performance}

The class-wise performance with $F1_\textnormal{frm}$ metric for CNN, RNN and CRNN methods along with the average and total duration of the classes are presented in Table~\ref{tab:class2}. CRNN outperforms both CNN and RNN on almost all classes. It should be kept in mind that each class is likely to appear together with different classes rather than isolated. Therefore the results in Table~\ref{tab:class2} present the performance of the methods for each class in a polyphonic setting, as would be the case in a real-life environment. The worst performing class for all three networks is cat meowing, which consists of short, harmonic sounds. We observed that cat meowing samples are mostly confused by baby crying, which has similar acoustic characteristics. Besides, short, non-impulsive sound events are more likely to be masked by another overlapping sound event, which makes their detection more challenging. CRNN performance is considerably better compared to CNN and RNN for gun shot, thunder, bird singing, baby crying and mixer sound events. However, it is hard to make any generalizations on the acoustic characteristics of these events that can explain the superior performance.

\subsubsection{Effects of filter shape}
The effect of the convolutional filter shape is presented in Table~\ref{tab:filt}. Since these experiments were part of the hyperparameter grid search, each experiment is conducted only once. Small kernels, such as (5,5) and (3,3), were found to perform the best in the experiments run on this dataset. This is consistent with the results presented in~\cite{sigtia2016end} on a similar task. The very low performance given for the filter shape (1,5) highlights the importance of including multiple frequency bands in the convolution when spectrogram based features are used as input for the CRNN.

\subsubsection{Number of parameters vs. accuracy}
The effect of number of parameters on the accuracy is investigated in Figure~\ref{fig:num_params}. The points in the figure represent the test accuracy with $F1_\textnormal{frm}$ metric for the hyperparameter grid search experiments. Each experiment is conducted one time only. Two observations can be made from the figure. For the same number of parameters, CRNN has a clear performance advantage over CNN and RNN. This indicates that the high performance of CRNN can be explained with the architectural advantage rather than the model size. In addition, there can be a significant performance shift for the same type of networks with the same number of parameters, which means that a careful grid search on hyperparameters (e.g. shallow with more hidden units per layer vs. deep with less hidden units per layer) is crucial in finding the optimal network structure.
 
\subsubsection{Frequency shift invariance}
\begin{figure}[!t]
\centering
%
\begin{tikzpicture}
\begin{axis}[
	width=\columnwidth-1.2cm,height=4.25cm,
	scaled x ticks=false,
    scale only axis,
    xticklabel style={/pgf/number format/fixed},
    grid,
    xtick={-2,-1,0,1,2},
    ytick={0,-1.0,-2.0,-3,-4,-5,-6,-7,-8},
    legend pos=north west,
    legend style={at={(axis cs: 1.3, 2.1), anchor=south east, inner sep=0pt, outer sep=0pt},draw=none, font=\scriptsize},
    x filter/.code=\pgfmathparse{#1},
	xlabel= Pitch shifted quartertones (24 quartertones per octave), ylabel=$\Delta \textnormal{F}1_\textnormal{frm}$(\%)]

\addplot+[red,mark options={fill=red},error bars/.cd,y dir=both,y explicit] coordinates {
(-2 ,-7.5) +- (1.1,1.1)
(-1 ,-5.8) +- (1.2,1.2)
(0 ,0) +- (0.9,0.9)
(1 ,-3.3) +- (1.2,1.2)
(2 ,-3.8) +- (1.3,1.3)   
};

\addplot+[green,mark options={fill=green},error bars/.cd,y dir=both,y explicit] coordinates {
(-2,-6.4) +- (0.5,0.5)
(-1,-5.9) +- (0.5,0.5)
(0,	0) +- (1.6,1.6)
(1,	-6.4) +- (0.9,0.9)
(2,	-6.9) +- (1.1,1.1)
};

\addplot+[blue,mark options={fill=blue},error bars/.cd,y dir=both,y explicit] coordinates {
(-2,-5.9) +- (1.2,1.2)
(-1,-3.7) +- (0.9,0.9)
(0,	0) +- (0.6,0.6)
(1,	-1.9) +- (0.9,0.9)
(2,	-2.4) +- (0.7,0.7)
};

\legend{CNN, RNN, CRNN}
\end{axis}
\end{tikzpicture}
\caption{Absolute accuracy change vs. pitch-shifting over $\pm$2 quartertones for CNN, RNN and CRNN.}
\label{fig:pitch_shift}
\end{figure}
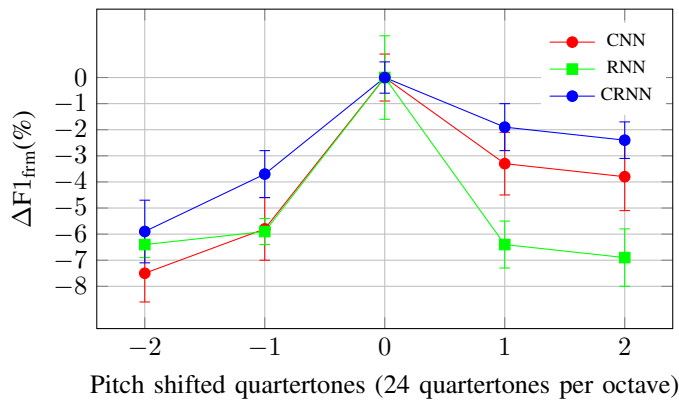

Sound events may exhibit small variations in their frequency content. In order to investigate the robustness of the networks to small frequency variations, pitch shift experiments are conducted and the absolute changes in frame-based F1 score are presented in Figure~\ref{fig:pitch_shift}. For these experiments, each network is first trained with the original training data. Then, using Librosa's pitch-shift function, the pitch for the mixtures in the test set is shifted by $\pm$2 quartertones. 
The test results show a significant absolute drop in accuracy for RNNs when the frequency content is shifted slightly. As expected, CNN and CRNN are more robust to small changes in frequency content due to the convolution and max-pooling operations. However, accuracy decrease difference between the methods diminishes for negative pitch shift, for which the reasons should be further investigated. It should be also noted that RNN has the lowest base accuracy, so it is relatively more affected for the same amount of absolute accuracy decrease (see Table~\ref{tab:general}).  

\begin{figure}[!t]
\includegraphics[width=\columnwidth]{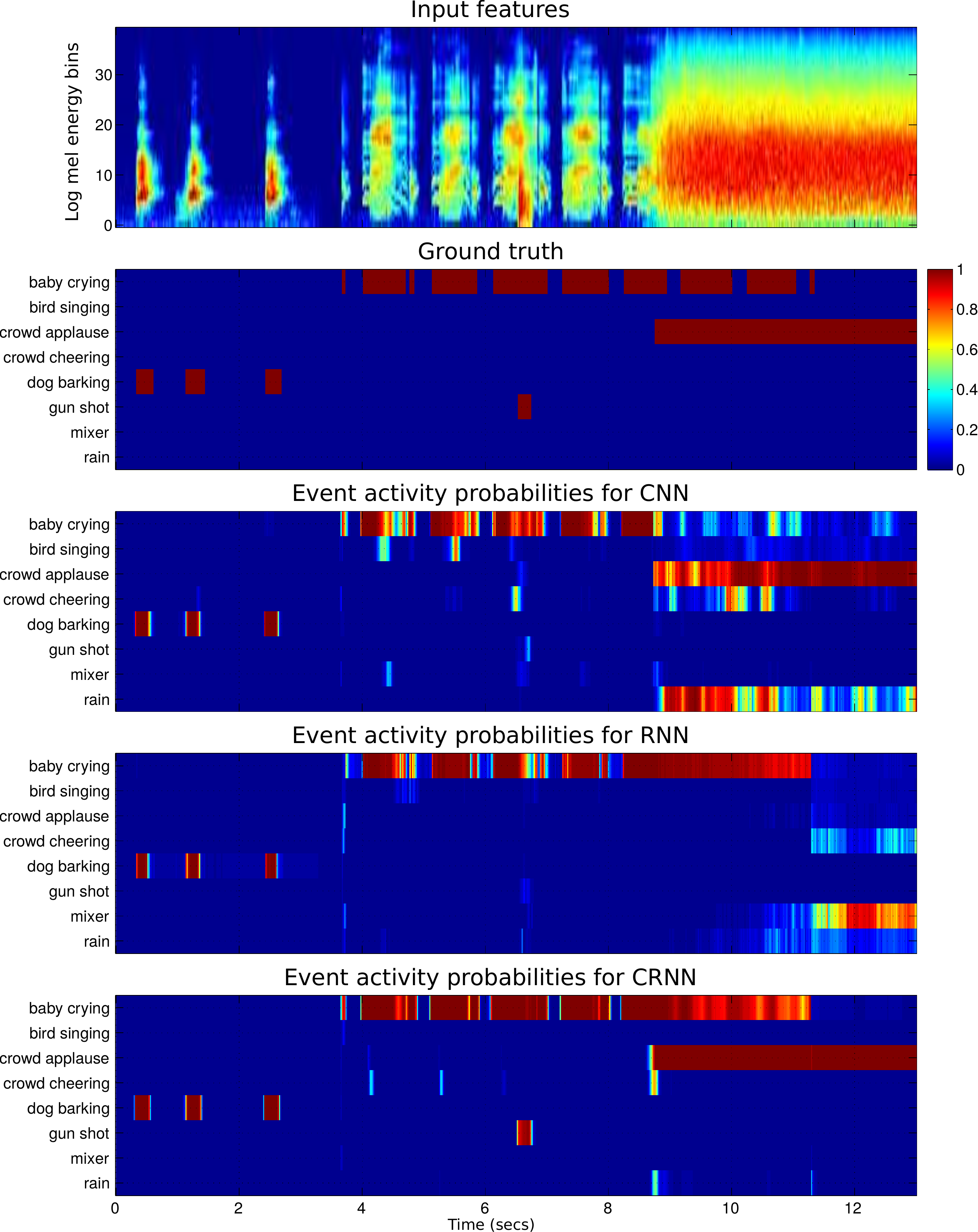}
\caption{Input features, ground truth and event activity probabilities for CNN, RNN and CRNN from a sequence of test examples from TUT-SED synthetic 2016.}

\label{fig:all_network_out}
\end{figure}

\subsubsection{Closer look on network outputs}
A comparative study on the neural network outputs, which are regarded as event activity probabilities, for a 13-second sequence of the test set is presented in Figure~\ref{fig:all_network_out}. For the parts of the sequence where \textit{dog barking} and \textit{baby crying} appear alone, all three networks successfully detect these events. However, when a \textit{gun shot} appears overlapping with \textit{baby crying}, only CRNN can detect the \textit{gun shot} although there is a significant change in the input feature content. This indicates the efficient modeling of the \textit{gun shot} by CRNN which improves the detection accuracy even in polyphonic conditions. Moreover, when \textit{crowd applause} begins to appear in the signal, it almost completely masks \textit{baby crying}, as it is evident from the input features. CNN correctly detects \textit{crowd applause}, but misses the masked \textit{baby crying} in this case, and RNN ignores the significant change in features and keeps detecting \textit{baby crying}. RNN's insensitivity to the input feature change can be explained with its input gate not passing through new inputs to recurrent layers. On the other hand, CRNN correctly detects both events and almost perfectly matches the ground truth along the whole sequence.

\subsection{TUT-SED 2009}
\label{subsec:casa}

For a comprehensive comparison, results with different methods applied to the same cross-validation setup and published over the years are shown in Table~\ref{tab:CASA}. The main metric used in these previous works is averaged over folds, and may be influenced by distribution of events in the folds (see Section~\ref{subsec:evalmet}). In order to allow a direct comparison, we have computed all metrics in the table the same way. 

First published systems were scene-dependent, where information about the scene is provided to the system and separate event models are trained for each scene  \cite{Mesaros2015_ICASSP, Heittola2013,Heittola2013b}. 
More recent work \cite{parascandolo2016recurrent,cakir2015}, as well as the current study, consist of scene-independent systems. 
Methods \cite{Heittola2013,Heittola2013b} are HMM based, using either multiple Viterbi decoding stages or NMF pre-processing to do polyphonic SED. 
In contrast, the use of NMF in \cite{Mesaros2015_ICASSP} does not build explicit class models, but performs coupled NMF of spectral representation and event activity annotations to build dictionaries. This method performs polyphonic SED through direct estimation of event activities using learned dictionaries.

The results on the dataset show significant improvement with the introduction of deep learning methods. CRNN has significantly higher performance than previous methods~\cite{Mesaros2015_ICASSP,Heittola2013,Heittola2013b,Mesaros2016_EUSIPCO}, and it still shows considerable improvement over other neural network approaches.




\begin{table}[!t]
\centering
\ra{1.1}
\caption{Results for TUT-SED 2009 based on the legacy F1. Methods marked with~$\star$ are trained in scene-dependent setting.}
\label{tab:CASA}
\resizebox{.9\columnwidth}{!}{%
\begin{tabular}{ @{} l *{1}{c}} 
\toprule
Method & Legacy $F1_{\textnormal{1sec}}$\\
\midrule
HMM multiple Viterbi decoding$^\star$~\cite{Heittola2013} & 20.4\\
NMF-HMM$^\star$~\cite{Heittola2013b} & 36.7 \\
NMF-HMM + stream elimination$^\star$~\cite{Heittola2013b} & 44.9\\
GMM$^\star$~\cite{Mesaros2016_EUSIPCO} & 34.6\\
\midrule
Coupled NMF$^\star$~\cite{Mesaros2015_ICASSP} & 57.8 \\
\midrule
FNN \cite{cakir2015} & 63.0\\
BLSTM \cite{parascandolo2016recurrent}    & 64.6\\
CNN   & 63.9$\pm$0.4\\
RNN    & 62.2$\pm$0.8\\
{\bf CRNN}  & {\bf 69.1$\pm$0.4}\\

\bottomrule
\end{tabular}
}
\end{table} 

\subsection{TUT-SED 2016}
The CRNN and RNN architectures obtain the best results in terms of framewise \textit{F1}. The CRNN outperforms all the other architectures for \textit{ER} framewise and on 1-second blocks. While the FNN obtains better results on the 1-second block \textit{F1}, this happens at the expense of a very large 1-second block \textit{ER}.

For all the analyzed architectures, the overall results on this dataset are quite low compared to the other datasets. This is most likely due the fact that TUT-SED 2016 is very small and the sounds events occur sparsely (i.e.\ a large portion of the data is silent). In fact, when we look at class-wise results (unfortunately not available due to space restrictions), we noticed a significant performance difference between the classes that are represented the most in the dataset (e.g. bird singing and car passing by, $F1_{\textnormal{frm}}$ around 50\%) and the least represented classes (e.g. cupboard and object snapping, $F1_{\textnormal{frm}}$ close to 0\%). Some other techniques might be applied to improve the accuracy of systems trained on such small datasets, e.g.\ training a network on a larger dataset and then retraining the output layer on the smaller dataset (transfer learning), or incorporating unlabeled data to the learning process (semi-supervised learning). 

\toni[inline]{These results do not look convincing. It seems that baseline methods are better in this particular setup. How come? Is the problem that there is not enough training material for CNN, RNN and CRNN based systems to be reliably trained in scene-dependent setting? My suggestion is that we omit these results from the paper, as they do look weak and will lower the impact of the paper in the eyes of the reviewers. We focus the paper to scene-independent systems only.}

\tv[inline]{I guess we don't have scene-independent results that could be shown? sharath was able to get better results than the baseline with his scene-dependent BLSTM. how come our scene-dependent systems are worse than his? this may be due to some additional technique such as class-specific threshold. our system being worse than baseline is not a good reason to drop the results (a good journal article would also show scenarios where the proposed system is not working so well). but if we include the results, it would be good to have an understanding why the results are not good.}

\toni[inline]{[NEW]side note: DCASE2016 dataset contains a lot events which are annotated, but not currently used. Two type of annotations are provided: 1) processed to contain only event classes which has good enough amount of examples and 2) raw annotations. Use of processed annotations, might confuse the network, as there is a lot additional event active outside annotations. Because of this dataset has quite low polyphony. To get such dataset working, one might need to help network to understand the data by creating "garbage"-class and have that active at target vector when non of the annotated events is active. 
}
\gb[inline]{very agreeable}
\emre[inline]{Tuomas, Heikki comments?}
\tv[inline]{overall the results are much worse than with the other datasets. this can due to smaller amount of material. and perhaps therfore simpler models (GMM / FNN) can therefore produce better results than CRNN? }

\tv[inline]{we probably should have included threshold C as one of the parameters to be optimized using the validation set. currently all the obtained ERs are above 1, which is worse than a system that does not output anything. therefore these results do not make much sense.}

\subsection{CHiME-Home}
The results obtained on CHiME-Home are reported in Table \ref{tab:chime}. For all of our three architectures there is a significant improvement over the previous results reported on the same dataset on the DCASE2016 challenge, setting new state-of-the-art results.

After the first series of experiments the CNN obtained slightly better results compared to the CRNN. The CRNN and CNN architecture used are almost identical, with the only exception of the last recurrent (GRU) layer in the CRNN being replaced by a fully connected layer followed by batch normalization. In order to test if the improvement in the results was due to the absence of recurrent connections or to the presence of batch normalization, we run again the same CNN experiments removing the normalization layer. As shown in the last row of \ref{tab:chime}, over 10 different random initializations the average EER increased to values above those obtained by the CRNN.



\begin{table}[!t]
\centering
\ra{1.1}
\caption{Equal error rate (EER) results for CHiME-Home development and evaluation datasets.}

\label{tab:chime} \normalsize
\resizebox{\columnwidth}{!}{%

\begin{tabular}{ @{} l *{3}{c}} 
\toprule
Method	& Development EER & Evaluation EER \\
\midrule
Lidy \textit{et al.}~\cite{Lidy2016}	&	17.8	&	16.6	\\
Cakir \textit{et al.}~\cite{Cakir2016dcase}	&	17.1	&	16.8	\\
Yun \textit{et al.}~\cite{Yun2016}		&	17.6	&	17.4	\\
\midrule
CNN					&	\textbf{12.6$\pm$0.5}&	\textbf{10.7$\pm$0.6}	\\
RNN					&	16.0$\pm$0.3&	13.8$\pm$0.4	\\
CRNN				&	\textbf{13.0$\pm$0.3}&	\textbf{11.3$\pm$0.6}	\\
\midrule
CNN (no batch norm)	&	15.1$\pm$1.7&	11.9$\pm$1.0	\\
\bottomrule
\end{tabular}

}
\end{table}

\subsection{Visualization of convolutional layers}
\begin{figure}[!t]
\centering
\includegraphics[width=0.48\textwidth]{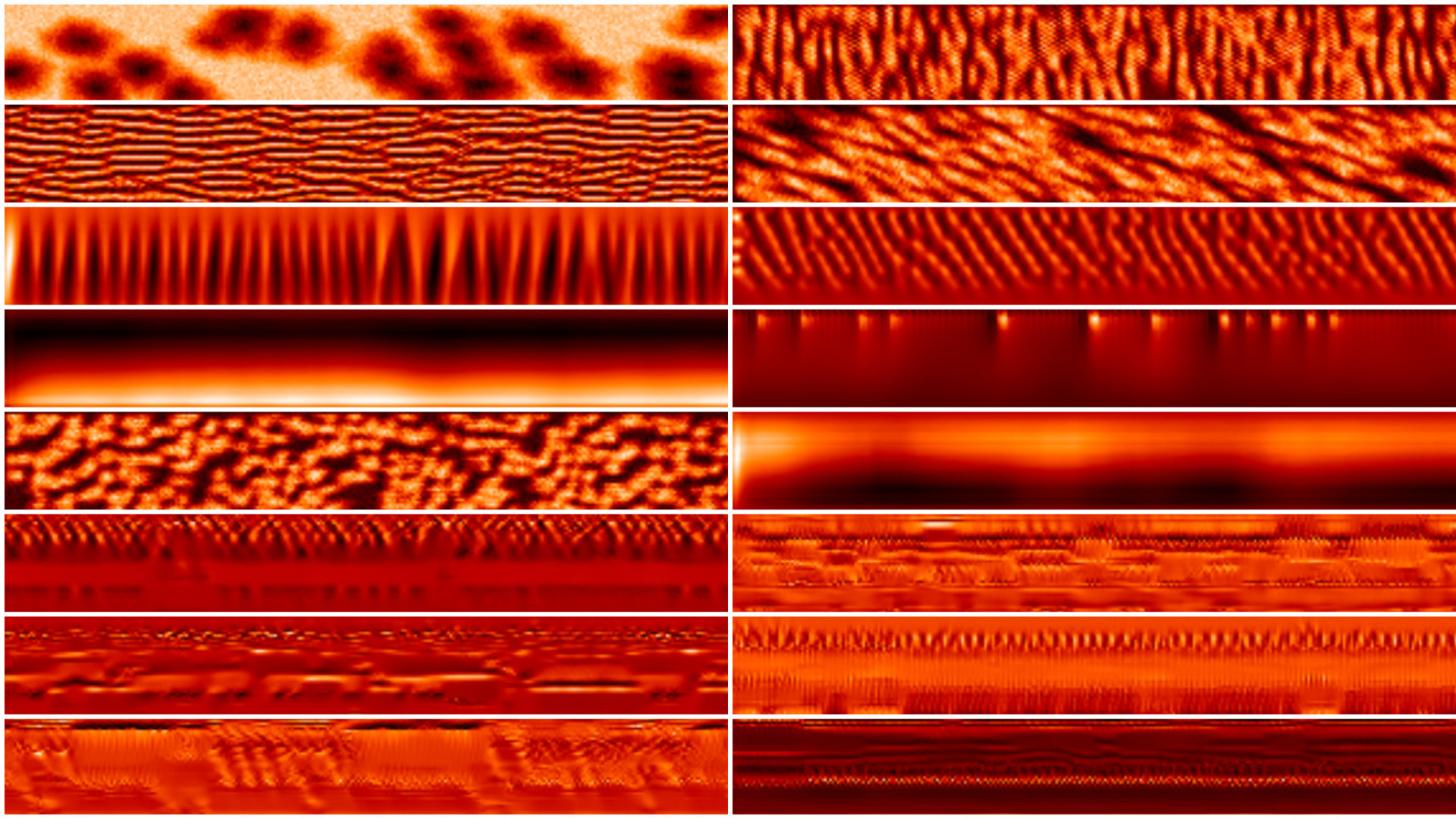}
\caption{Two columns of crops from input patterns that would strongly activate certain neurons from different layers of the CRNN. On the horizontal axis is time, on the vertical axis mel bands. On both columns the rows 1 and 2 are from neurons in the first convolutional layer, rows 3 to 5 from the second, and rows from 6 to 8 from the third.}
\label{fig:filts}
\end{figure}

Here we take a peek at the representation learned by the networks. More specifically, we use the technique described in \cite{simonyan2013deep} to visualize what kind of patterns in the input data different neurons in the convolutional layers are looking for. We feed the network a random input whose entries are independently drawn from a Gaussian distribution with zero mean and unit variance. We choose one neuron in a convolutional layer, compute the gradient of its activation with respect to the input, and iteratively update the input through gradient ascent in order to increase the activation of the neuron. If the gradient ascent optimization does not get stuck into a weak local maximum, after several updates the resulting input will strongly activate the neuron. We run the experiment for several convolutional neurons in the CRNN networks trained on TUT-SED Synthetic 2016 and TUT-SED 2009, halting the optimization after 100 updates. In Figure \ref{fig:filts} we present a few of these inputs for several neurons at different depth. The figure confirms that the convolutional filters have specialized into finding specific patterns in the input. In addition, the complexity of the patterns looked for by the filters seems to increase as the layers become deeper.

\section{Conclusions}
\label{sec:concl}
In this work, we proposed to apply a CRNN---a combination of CNN and RNN, two complementary classification methods---on a polyphonic SED task. The proposed method first extracts higher level features through multiple convolutional layers (with small filters spanning both time and frequency) and pooling in frequency domain; these features are then fed to recurrent layers, whose features in turn are used to obtain event activity probabilities through a feedforward fully connected layer. In CRNN, CNN's capability to learn local translation invariant filters and RNN's capability to model short and long term temporal dependencies are gathered in a single classifier. The evaluation results over four datasets show a clear performance improvement for the proposed CRNN method compared to CNN, RNN, and other established methods in polyphonic SED. 

Despite the improvement in performance, we identify a limitation to this method. As presented in TUT-SED 2016 results in Table~\ref{tab:general}, the performance of the proposed CRNN (and of the other deep learning based methods) strongly depends on the amount of available annotated data. TUT-SED 2016 dataset consists of 78 minutes of audio of which only about 49 minutes are annotated with at least one of the classes. When the performance of CRNN for TUT-SED 2016 is compared to the performance on TUT-SED 2009 (1133 minutes) and TUT-SED Synthetic 2016 (566 minutes), there is a clear performance drop both in the absolute performance and in the relative improvement with respect to other methods. Dependency on large amounts of data is a common limitation of current deep learning methods. 



The results we observed in this work, and in many other classification tasks in various domains, prove that deep learning is definitely worth further investigation on polyphonic SED. As a future work, semi-supervised training methods can be investigated to overcome the limitation imposed by small datasets. Transfer learning~\cite{bengio2012deep,yosinski2014transferable} could be potentially applied with success in this setting: by first training a CRNN on a large dataset (such as TUT-SED Synthetic 2016), the last feedforward layer can then be replaced with random weights and the network fine-tuned on the smaller dataset. 

Another issue worth investigating would be a detailed study over the activations from different stages of the proposed CRNN method. For instance, a class-wise study over the higher level features extracted from the convolutional layers might give an insight on the common features of different sound events. Finally, recurrent layer activations may be informative on the degree of relevance of the temporal context information for various sound events.

\ifCLASSOPTIONcaptionsoff
  \newpage
\fi



%


\bibliographystyle{IEEEtran}
\bibliography{refs}

%


\begin{IEEEbiography}[{\includegraphics[width=1in,height=1.2in,clip,keepaspectratio]{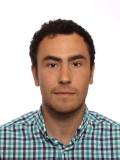}}]{Emre Cakir}
received his BSc. degree in electrical-electronics engineering from Middle East Technical University, Ankara, Turkey in 2013 and the MSc. degree in Information Technology from Tampere University of Technology (TUT), Finland in 2015. He has been with the Audio Research Group in TUT since February, 2014 where he currently continues his PhD studies. His main research interests are sound event detection in real-life environments and deep learning. 
\end{IEEEbiography}


\begin{IEEEbiography}[{\includegraphics[width=1in,height=1.2in,clip,keepaspectratio]{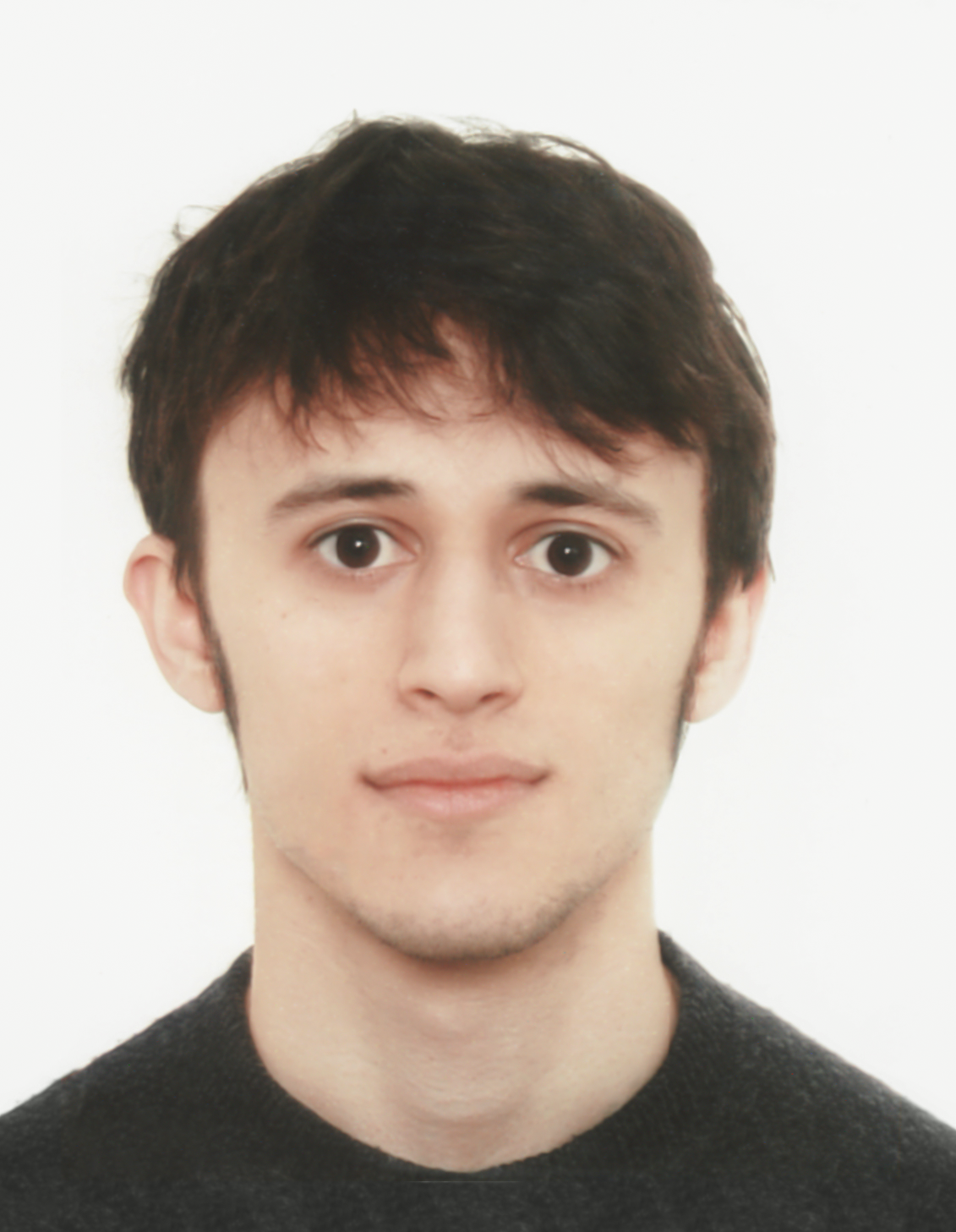}}]{Giambattista Parascandolo}
received his B.Sc.\ degree from the Department of Mathematics at University of Rome Tor Vergata, Italy, in 2013, and his M.Sc.\ degree in Information Technology from Tampere University of Technology (TUT), Finland, in 2015. He is a Project Researcher at the Audio Research Group in TUT, where he has been since February, 2015. His main research interests are deep learning and machine learning.
\end{IEEEbiography}



\begin{IEEEbiography}[{\includegraphics[width=1in,height=1.2in,clip,keepaspectratio]{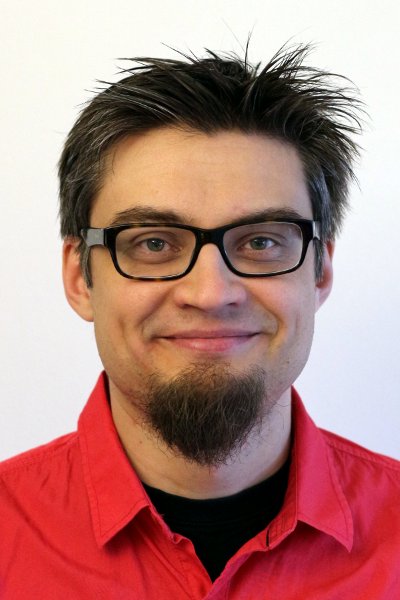}}]{Toni Heittola} received his M.Sc. degree in Information Technology from Tampere University of Technology (TUT), Finland, in 2004. He is currently pursuing the Ph.D. degree at TUT. His main research interests are sound event detection in real-life environments, sound scene classification and audio content analysis.
\end{IEEEbiography}

\begin{IEEEbiography}[{\includegraphics[width=1in,height=1.2in,clip,keepaspectratio]{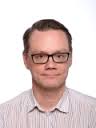}}]{Heikki Huttunen}
received his Ph.D. degree in Signal Processing at Tampere University of Technology (TUT), Finland, in 1999. Currently he is a university lecturer at the Department of Signal Processing at TUT. He is an author of over 100 research articles on signal and image processing and analysis. His research interests include Optical Character Recognition, Deep learning, and Pattern recognition and Statistics.
\end{IEEEbiography}

\begin{IEEEbiography}[{\includegraphics[width=1in,height=1.2in,clip,keepaspectratio]{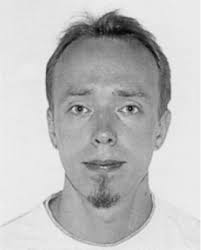}}]{Tuomas Virtanen}
is an Academy Research Fellow and an adjunct professor at Department of Signal Processing, Tampere University of Technology (TUT), Finland. He received the M.Sc. and Doctor of Science degrees in information technology from TUT in 2001 and 2006, respectively. He is known for his pioneering work on single-channel sound source separation using non-negative matrix factorization based techniques, and their application to noise-robust speech recognition, music content analysis and audio event detection. In addition to the above topics, his research interests include content analysis of audio signals in general and machine learning.He has received the IEEE Signal Processing Society 2012 best paper award.
\end{IEEEbiography}



\end{document}